\documentclass[runningheads]{llncs}
\usepackage[T1]{fontenc}
\usepackage{graphicx}
\usepackage{csquotes}
\usepackage{xcolor}
\usepackage{hyperref}
\newcommand{\spaceunderfig}{-0.2cm}
\begin{document}
\title{Procedural Knowledge Ontology (PKO)}
%
%
\author{Valentina Anita Carriero \orcidID{0000-0003-1427-3723} \and
\\Mario Scrocca \orcidID{0000-0002-8235-7331} \and 
Ilaria Baroni \orcidID{0000-0001-5791-8427} \and 
\\Antonia Azzini \orcidID{0000-0002-9066-1229} \and 
Irene Celino \orcidID{0000-0001-9962-7193}}
\authorrunning{V.A. Carriero et al.}
%
\institute{Cefriel -- Politecnico di Milano, viale Sarca 226, 20126 Milano, Italy\\
\email{name.surname@cefriel.com}}
\maketitle              
\begin{abstract}
Processes, workflows and guidelines are core to ensure the correct functioning of industrial companies: for the successful operations of factory lines, machinery or services, often industry operators rely on their past experience and know-how. The effect is that this \textit{Procedural Knowledge} (PK) remains tacit and, as such, difficult to exploit efficiently and effectively.
This paper presents PKO, the Procedural Knowledge Ontology, which enables the explicit modeling of procedures and their executions, by reusing and extending existing ontologies. 
PKO is built on requirements collected from three heterogeneous industrial use cases and can be exploited by any AI and data-driven tools that rely on a shared and interoperable representation to support the governance of PK throughout its life cycle.
We describe its structure and design methodology, and outline its relevance, quality, and impact by discussing applications leveraging PKO for PK elicitation and exploitation.


\end{abstract}

\section{Introduction}\label{sec:introduction}

Processes, workflows and guidelines make it possible to ensure the correct functioning of industrial companies. However, for the successful operations of factory lines, machinery or services, it is often the case that industry operators rely on their past experience and know-how. The downside is that this \textit{Procedural Knowledge} (PK) remains implicit and, as such, difficult to exploit in an efficient and effective manner.
Even when it is documented, the only digital source of PK is expressed by means of natural language, and is spread across different documents and systems. 
Having PK expressed explicitly and in a structured and semantic format, thanks to an ontology, would guarantee interoperability between different data sources, enable reasoning and inference, and support PK reuse.
In an industrial scenario, it would make it easier to access and apply PK when it comes to procedures execution. Moreover, such PK management could result in (i) improved compliance with industrial processes and standards, limiting undesired errors when executing the procedures, with a potential positive impact on business objectives, functioning of industrial services and systems, costs, and employees' safety; and (ii) additional support to knowledge transfer and (re)training/onboarding of (new) employees.

This paper presents PKO\footnote{\url{https://w3id.org/pko}}, the Procedural Knowledge Ontology, which explicitly models procedures, their executions, and related resources, by reusing and extending existing  ontologies. 
PKO is built on requirements collected from three heterogeneous industrial use cases~\cite{carriero2024towardspko}. 
It can be used by any AI and data-driven solutions that needs to rely on a shared and interoperable representation that can support the governance of PK throughout its life cycle.

The remainder of this paper is organized as follows. Section \ref{sec:related} discusses relevant standards and ontologies in the procedural domain. Section \ref{sec:methodology} briefly summarises how we applied the ontology engineering methodology to the development of PKO. In Section \ref{sec:requirements} we present the collected requirements, and the process followed to analyse the domain. Section \ref{sec:onto} illustrates the Procedural Knowledge Ontology, with its two main modules, and the reused ontologies. Section \ref{sec:howto} exemplifies PKO intended usage. Section \ref{sec:evaluation-impact} provides an evaluation of PKO based on automatic pitfalls detection and ontology evaluation criteria, and discusses its potential impact by discussing applications leveraging PKO for PK elicitation and exploitation. Section \ref{sec:publication} reports on PKO availability and sustainability. Finally, Section \ref{sec:conclusions} draws the conclusions.

\section{Related Work}\label{sec:related}
\paragraph{(Business) Process Modelling} In the business process (BP) community, processes are modeled as collections of activities~\cite{harth2024process}, as in the Business Process Model and Notation (BPMN)\footnote{\url{https://www.omg.org/spec/BPMN}}. BPMN is a widely adopted visual language for representing processes. It supports the representation of activities, events, gateways, and data requirements (like input/output data). In BPMN, the term \emph{activity} is used to denote a piece of work within a process, i.e., a step, and is further specialised in \emph{task} (an atomic unit of work) and \emph{subprocess} (a group of multiple tasks). \textcolor{black}{So, in BPMN a \emph{task}, by being a subclass of \emph{activity}, is an activity itself, defined at process specification level.} Instead, an \emph{event} represents something that happens during the course of a process, possibly triggering a certain activity. \emph{States} are not explicitly addressed, however, a change of state may be represented using the \emph{none event}, a specific type of event without a trigger condition. While BPMN focuses on a very detailed representation of the process specification, it does not make any distinction between process specification and process execution~\cite{annane2020comparing}.
Accordingly, ontologies that start from existing BP notations and languages, and aim to represent (a part of) their constraints -- like the Business Process Modeling Ontology (BPMO)\footnote{Currently, the ontology is not available online.}~\cite{cabral2009business} and the BPMN Based Ontology (BBO)\footnote{\url{https://www.irit.fr/recherches/MELODI/ontologies/BBO}}~\cite{annane2019bbo} -- either only model the process specification, excluding the process execution from their scope, or they mix between execution and specification, without a clear distinction. Moreover, support for organization and resource related attributes is very limited~\cite{harth2024process}.
A different approach is applied by Petri-Net~\cite{murata1989petri}, a mathematical and graphical language for modelling workflows. Petri-Net considers two types of nodes, \emph{places}, which represent \emph{states}, and \emph{transitions}, which represent events and \emph{activities}/\emph{actions} that change the values of the states. However, time constraints, resources and agents are not supported. Some work has been done on translating such language into ontologies (e.g.,~\cite{gavsevic2006petri}), but no ontology could be found online.
The Process Specification Language (PSL)~\cite{pouchard2005iso} is an ISO standard created at the National Institute of Standards and Technology (NIST) for representing manufacturing processes. 
PSL makes a distinction between an \emph{activity} (intended as a specification, similarly to BPMN) and an \emph{activity occurrence}, i.e. its execution. However, it is not possible to define process specifications independently from process executions, since the necessary constraints (e.g., sequential order) can be applied only to instances of activity occurrences~\cite{annane2020comparing}. The PSL ontology\footnote{\url{https://web.archive.org/web/20071230103337/http://www.mel.nist.gov/psl/ontology.html}} is formalised as a set of axioms written in CLIF\footnote{Common Logic Interchange Format}.

\paragraph{Provenance-related ontologies} Relevant work on modelling plans and their executions includes provenance-related ontologies. 
The Provenance Ontology (PROV-O)\footnote{\url{https://www.w3.org/TR/prov-o/}} models provenance information generated in different systems/contexts 
and has been developed both for direct reuse and as a reference model for creating domain-specific provenance ontologies. Its core class is \texttt{Activity}, however, unlike BPMN, an \emph{activity} is defined as ``something that occurs over a period of time and acts upon or with entities'', thus it is something that happens, and can be associated with agents, time information, 
and other entities, e.g., a resource generated by the activity. 
Similarly, in the Enterprise Ontology (EO)~\cite{uschold1998enterprise}, aimed at representing executable process models to help users perform their tasks, an activity is an execution of an \emph{activity specification}.
In PROV-O, an activity may also be linked to a \emph{plan}, that is ``a set of actions or steps intended by one or more agents to achieve some goals'', and has been adopted by an agent in association with a certain activity.
However, PROV-O does not model the plan further, nor its execution.
The P-Plan ontology\footnote{\url{http://purl.org/net/p-plan}} extends PROV-O with classes and properties to describe plans and plan executions. The \emph{plan} is linked to its \emph{steps}, which can be aligned with the BPMN concept of activity intended as a unit of work within a process, while by \emph{activity} P-Plan means the execution of a step by an \emph{agent}. 
The Open Provenance Model for Workflows (OPMW)\footnote{\url{https://www.opmw.org/model/OPMW/}}~\cite{garijo2011new} is an ontology for describing workflow traces and their template in scientific processes, and reuses both PROV-O and P-Plan. It represents plans (called \texttt{WorkflowTemplate}s), steps (called \texttt{WorkflowTemplateProcess}es), and their executions (called \texttt{WorkflowExecutionProcess}es).
OPMW introduces the concept of \emph{workflow execution account}, which is a specialisation of \emph{bundle} from PROV-O (i.e., a set of provenance descriptions), and allows to represent the execution view from the perspective of the system, tracking also its \emph{status} (e.g., \emph{success}).
However, P-Plan and OPMW do not capture complex control-flow constructs as BPMN-based ontologies, but steps are only linked with properties representing the sequence. 
Also, \emph{states} within plans are not explicitly modeled.

\paragraph{Foundational ontologies} Foundational ontologies aim at providing domain-\-in\-de\-pend\-ent general classes and properties, that can be used as a common ground to build domain-specific ontologies. 
Dolce+DnS Ultralite (DUL)\footnote{\url{http://www.ontologydesignpatterns.org/ont/dul/DUL.owl}}, which derives from the Descriptive Ontology for Linguistic and Cognitive Engineering (DOLCE)\footnote{\url{https://www.loa.istc.cnr.it/index.php/dolce/}}~\cite{borgo2022dolce}, includes also an extension about plans\footnote{\url{http://www.ontologydesignpatterns.org/ont/dul/PlansLite.owl}}.
Here, a \emph{plan} defines a set of \emph{tasks} to be executed, that can be of different types (e.g., elementary task, complex task, etc.). In this case, the term task can be aligned with \emph{step} from P-Plan and \emph{activity} from BPMN. The \emph{plan execution} is also modelled, and an \emph{action} executes a certain task defined in the modeled, so it has a similar semantics to \emph{activity} from P-Plan.
\emph{States} are not explicitly addressed, but they can be modelled as events (e.g., ``what has brought a certain state to occur'').
The Basic Formal Ontology (BFO)~\cite{otte2022bfo} defines the concept \emph{process} as an ``occurrent that has some temporal proper part and for some time has some material entity as participant'', thus the process is defined as an event, something that happens, as opposed to \emph{plan} from DUL, PROV-O and BPMN.
Schema.org\footnote{\url{https://schema.org/}}, aimed at enabling structured data markup on web pages, models special kinds of procedures with the \texttt{HowTo} class, that includes sequentially ordered \texttt{HowToStep}s. However, the execution and related resources are not taken into account.


\section{Applied Methodology for Ontology Engineering}\label{sec:methodology}

To develop the ontology, we relied on the Linked Open Terms (LOT) methodology\footnote{\url{https://lot.linkeddata.es/}}~\cite{poveda22-lot}, a lightweight method for developing ontologies and vocabularies related to the industry domain.
It defines iterations over a basic workflow of the following main activities: (i) ontology requirements specification, (ii) ontology implementation, (iii) ontology publication, and (iv) ontology maintenance.

\paragraph{Ontology requirements specification}
The ontology requirements specification activity refers to the process of defining the goal and scope of the ontology and specifying the requirements that the ontology should satisfy.
We identified a list of use cases and user stories that describe different scenarios that need to be supported by the data and modelled in the vocabulary, and a set of documents and resources relevant to the domain.
Based on this, we produced a set of functional requirements in the form of \emph{competency questions} (CQs), that are  questions to be answered by the ontology \cite{gruninger95cqs}, and \emph{facts}, natural language sentences describing the  entities and terminology (e.g., attributes specifying a certain term to be used in the ontology).
In this activity, the collaboration between ontology engineers, users and domain experts was central to ensure that the collected requirements were valid and complete.
All collected and validated requirements have been properly documented. Details are included in Section \ref{sec:requirements}.

\paragraph{Ontology implementation}
Before building the ontology using a formal language, we designed an initial conceptual model informally representing the domain of knowledge, with a proposal of concepts, relationships and their hierarchies. Then, we encoded and extended the conceptual model in the OWL ontology language. Following a recommended practice in ontology engineering, we reused existing ontological resources that partially covered some of our modelling problems, to improve interoperability \cite{carriero2020landscape}. 
Details can be found in Section \ref{sec:onto}.

\paragraph{Ontology publication} 
The aim of this activity is to make the ontology accessible and documented online.
After selecting a suitable licence, we published a first release of the ontology in different syntaxes (Turtle, RDF/XML, JSON-LD, N-Triples), following a content-negotiation mechanism that makes it available in both human- and machine-readable format, following the best practices for vocabularies on the web provided by the W3C\footnote{\url{http://www.w3.org/TR/swbp-vocab-pub/}}.
The documentation includes (i) a description of the ontology and its ontology entities, (ii) relevant metadata (e.g., title, creator, date of creation), (iii) a graphical representation of the ontology, and (iv) the list of requirements addressed. Usage examples will be included to help reuse.
More details are discussed in Section \ref{sec:publication}.

\paragraph{Ontology maintenance}
The published ontology should be maintained available and up-to-date, addressing possible bugs, suggestions, improvements, and new emerging requirements. In our case, for now, we simply set up a mechanism to facilitate the collection of suggestions and requests from the ontology adopters.
Section \ref{sec:publication} discusses maintenance, sustainability and future evolutions of PKO.


\section{Ontology Requirements}\label{sec:requirements}

We performed a sequence of hands-on workshops to collect the necessary input to develop the ontology from domain experts and expected users of three different industrial scenarios (safety procedures in a plant, CNC commissioning processes, and mixed human-machine activities in grid management, cf.~\cite{carriero2024towardspko}). 
Specifically, each use case defined a set of capabilities, representing the features that a Procedural Knowledge Management System (PKMS)  should address. Such capabilities were collected based on the definition of \emph{as-is} and \emph{to-be} scenarios. The list of capabilities was used as a starting point to outline the initial set of use cases and user stories that would guide the ontology development (\emph{Use
case specification} activity). 
Moreover, we discussed with experts also the available documentation relevant to the domain of knowledge (\emph{Data exchange identification} activity). Such documentation includes procedures in the form of text and tables in various formats (e.g., PDF, Excel), manuals and guidelines related to procedures, images, videos, examples of procedure executions. 

As a result, we were able to define the purpose and scope of the ontology (\emph{Purpose and scope identification} activity). 
Then, from the collected capabilities we derived the functional ontology requirements, as \emph{competency questions} (CQ) and \emph{facts} (\emph{Functional ontological requirements proposal} activity). 
Each CQ was discussed with, and validated by, our domain experts (\emph{Functional ontological requirements completion} activity).
Facts and CQs that we defined and documented so far represent our  ontological requirements for procedural knowledge representation (\emph{ORSD formalization} activity).

\paragraph{Collected ontology requirements} Our initial set of requirements includes 54 competency questions and 67 facts. All requirements can be found on GitHub\footnote{\url{https://github.com/perks-project/pk-ontology/tree/master/requirements}}. 
Future iterations may lead to the definition of additional CQs and facts.
Specifically, a subset of CQs focuses on the \emph{specification} of the procedure, i.e., how the procedure is composed (its steps and their sequential order), possible versions of the procedure (and their order), how the procedure was created, when, and by whom, etc. Another subset is focused on the \emph{execution} of the procedure, i.e., when and by whom a procedure has been executed, possible errors/problems arisen during execution, which steps of a procedure have been actually executed, etc. Additional CQs revolve around the agents involved in the creation/execution of the procedure, and their roles, and around relevant documents/images/videos/FAQs relevant to the procedure, with their metadata.

Let us make an example of how a feature of the PKMS, defined in collaboration by domain experts, users and software/ontology engineers, translates into competency questions and facts relevant for developing the ontology.

\begin{quotation}
\textbf{PKMS requirement.} \enquote{A procedure may have different steps and/or sub-procedures and may refer to other data sources of the company (e.g., documents/image/video) associated with it.}

\textbf{Competency Questions.} Two CQs  for this requirement  are: \enquote{Which are the steps of a procedure?}, and \enquote{From which resource was a piece of information collected and associated to a procedure?}.

\textbf{Facts.} Two facts that correspond to the CQs are: \enquote{A step of a procedure can be either atomic, or decomposed into a subset of multiple steps}, and \enquote{A step can refer to one or more resources (e.g., documents)}.
\end{quotation}

\section{Procedural Knowledge Ontology (PKO)}\label{sec:onto}
Based on the collected requirements, as a first step towards the ontology implementation, we identified the main concepts of the ontology~\cite{carriero2024towardspko}. 
We clustered the requirements to obtain a conceptual representation of the domain into separate coherent subdomains based on conceptual areas. 
We came up with six conceptual areas that cluster our concepts, each having a representative concept that we use for naming it: \emph{Procedure}, \emph{Step}, \emph{Change of Procedure Status}, \emph{Procedure Execution}, \emph{Resource}, \emph{Agent}.
As a second step, we enriched this high-level overview of concepts with more granular, and possibly hierarchically related, concepts, relations and attributes (\emph{Ontology conceptualization}), using the UML-like template provided by the drawio.com diagramming tool.

While performing this activity, we realised that a subset of requirements was not specifically related to the procedural domain, but to the industrial domain in general. For this reason, we decided to include the modelling of such requirements (6 CQs and 4 facts, so far\footnote{\url{https://github.com/perks-project/pk-ontology/blob/master/requirements/ontology-requirements-industry.xlsx}})
in a dedicated ontology module, in order to make it more easily reusable. 
In future evolutions of PKO, we may decide to further split it in additional modules to better structure the content.

After reaching a stable first version of the conceptual model, we started formalising it into the OWL ontology language. As a first step, we looked for existing ontologies that could possibly already address our requirements (\emph{Ontology reuse} activity). We both searched for scientific papers on published ontologies, and relied on existing catalogues of general purpose ontologies, like Linked Open Vocabularies (LOV)\footnote{\url{https://lov.linkeddata.es}}, and catalogues of domain-specific ontologies, like the Industry Portal\footnote{\url{https://industryportal.enit.fr/}}.
We apply a direct \emph{soft} reuse, i.e. we include selected ontology terms (their original URIs) in the ontology under development \cite{carriero2020landscape}. 
The ontologies selected for reuse so far are:
\begin{itemize}
    \item \emph{PROV-O}, for general concepts for modelling agents, activities, and provenance-related information
    \item \emph{P-Plan}, which extends PROV-O for modelling plans, with their steps, and plans executions
    \item \emph{D-CAT}\footnote{Data Catalog Vocabulary, cf. \url{https://www.w3.org/TR/vocab-dcat-3/}}, for representing metadata about different kinds of resources,
    \item \emph{DCMI Metadata Terms}\footnote{\url{https://www.dublincore.org/specifications/dublin-core/dcmi-terms/}}, a general purpose metadata vocabulary for describing resources of any type,
    \item \emph{Time ontology}\footnote{Cf. \url{https://www.w3.org/TR/owl-time/}}, an ontology of temporal concepts
    \item \emph{ADMS}\footnote{Asset Description Metadata Schema, cf. \url{https://www.w3.org/TR/vocab-adms/}}, a profile of DCAT for describing (semantic) assets with metadata, like file, licence, repository, etc.
    \item \emph{Metadata4Ing}\footnote{\url{http://w3id.org/nfdi4ing/metadata4ing\#}}, an ontology for representing research data and the whole data generation process, including tools
    \item \emph{PRO}\footnote{Publishing Roles Ontology, cf. \url{http://www.sparontologies.net/ontologies/pro}}, for describing roles of agents in time
\end{itemize}

BPMN-based models have been discarded since they do not address the procedure execution level, nor the modelling of organizational and resource related metadata. 
\textcolor{black}{PKO represents a procedure as something that can be generated and executed by a human and does not need to be directly executed by a machine, with a \enquote{general/high-level} definition of sequences of steps. Conversely, BPMN is intended for \enquote{more complex/fine-grained} representation of executable processes, thus needing complex control-flow constructs, which is not needed in our use cases (see Section \ref{sec:related}).}
Finally, while the state of the procedure (e.g., draft, archived) emerged as a relevant requirement, along with the (change of) state of the procedure execution, the (change of) state of the machine/system while executing the procedure was not part of our initial set of requirements.

\begin{figure}[h]
\centering
\includegraphics[width=\linewidth]{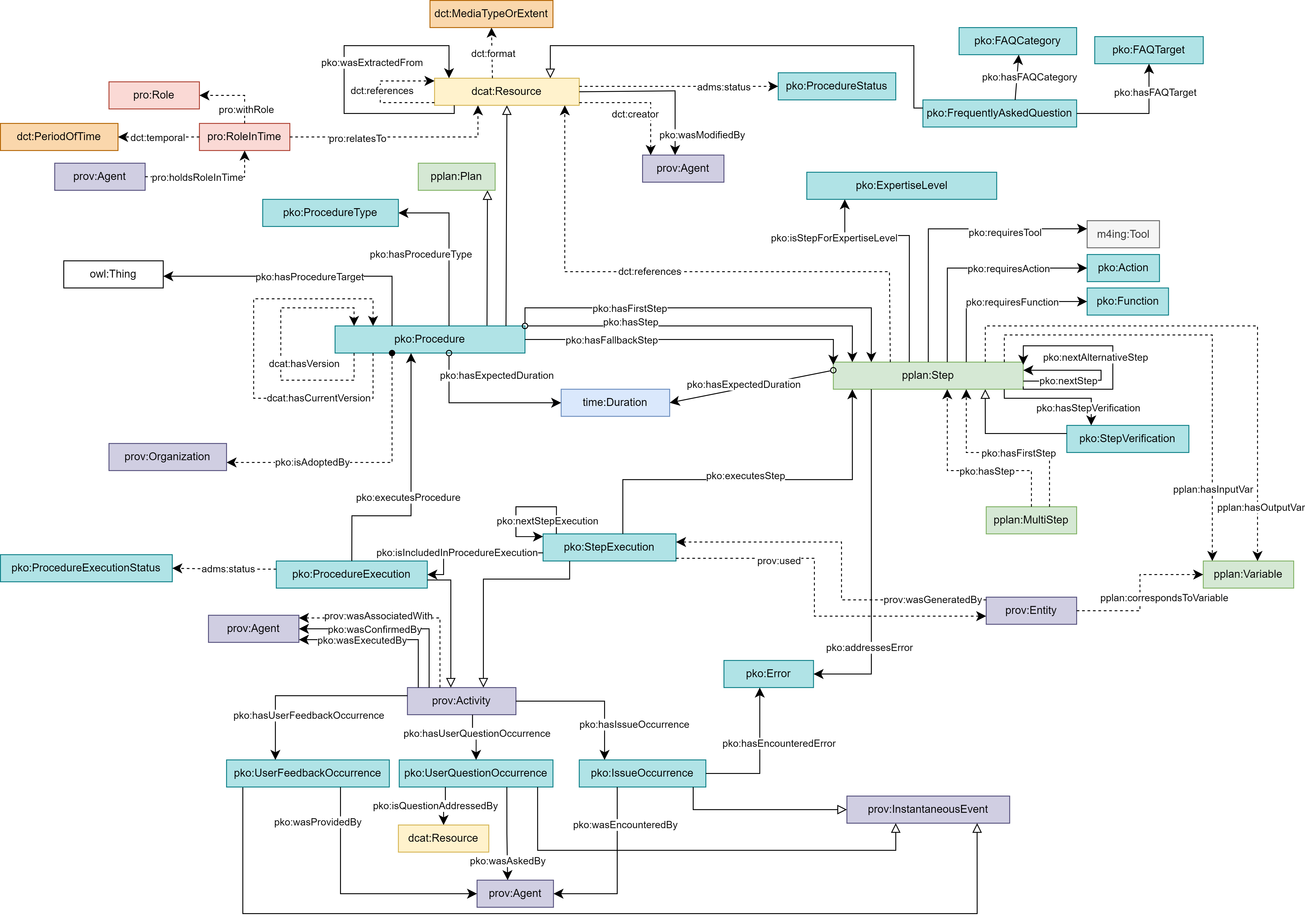}
\caption{Graphical representation of PKO core.}
\label{fig:pko-short}
\vspace{\spaceunderfig}
\end{figure}

We encoded the ontology (both PKO \emph{core}\footnote{\url{https://github.com/perks-project/pk-ontology/blob/master/ontology/pko.ttl}} and PKO \emph{industry}\footnote{\url{https://github.com/perks-project/pk-ontology/blob/master/ontology/pko-ind.ttl}}) in OWL (\emph{Ontology encoding} activity) using the Chowlk tool\footnote{\url{https://chowlk.linkeddata.es/}} \cite{chavez2021converting}, which is an online converter that transforms XML ontology diagrams, generated with draw.io using the Chowl notation, into OWL code. Comments and notes of usage that clarify the meaning and use of a class/property\footnote{Using the annotation property \texttt{skos:scopeNote}, cf. \url{https://www.w3.org/2012/09/odrl/semantic/draft/doco/skos\_scopeNote.html}} have been encoded in separate files.
Figure \ref{fig:pko-short} depicts all classes and relations of the PKO core ontology, except for datatype properties\footnote{The complete graphical representation is available on GitHub.}.

\paragraph{PKO core} The main class of the ontology is \texttt{Procedure}, which is defined as a sequence of actions to be executed in order to achieve a desired outcome. Such outcome can be expressed in terms of a \texttt{ProcedureType} (e.g., a maintenance activity), that is the type of procedure being executed on a procedure target (e.g., a production line for a specific type of machines). The class Procedure is declared as a subclass of both \texttt{Plan}, and \texttt{Resource} (intended as any entity that can be included in a catalog.)
If someone wants to manage version control, and keep track of updates and revisions to the same procedure, it is possible to link an \enquote{abstract} procedure to its versioned procedures (\texttt{hasVersion}), which are in a sequential order (\texttt{next/previousVersion}). 
A Procedure can be associated with a \texttt{ProcedureStatus} (e.g., draft, approved). 
It is linked to all its \texttt{Step}s with the property \texttt{hasStep}. A \texttt{Step} groups one or more \texttt{Action}s (from a human -- \texttt{requiresAction}) or \texttt{Function}s (from an algorithm/software -- \texttt{requiresFunction}) to execute a portion of a \texttt{Procedure}, possibly with the use of some \texttt{Tool}s (\texttt{requiresTool}). 

A \texttt{Step} can be decomposed as a set of multiple steps itself (\texttt{Multistep}). The class \texttt{StepVerification} represents the way in which the execution of a \texttt{Step} can be verified.

When different \texttt{Agent}s (e.g., a junior vs a senior technician) perform the same procedure, we may need to specify the \texttt{ExpertiseLevel} of each (Multi)Steps, meaning that the step is targeted at an agent with that level of expertise. This allows for different formulations and levels of detail of the same steps.

A certain \texttt{Procedure} can be executed one or multiple times, by one or more agents, at a certain time. This is represented by the classes of \texttt{ProcedureExe\-cu\-tion} and \texttt{StepExecution}, which are defined as subclasses of \texttt{Activity}, and thanks to which we can track the execution of the procedure as a whole, and as a composition of executions of the individual steps. 
The \texttt{ProcedureExecution} can be associated with a \texttt{ProcedureExecutionSta\-tus} (e.g., in progress, completed).

The \texttt{UserFeedbackOccurrence} represents the case in which an \texttt{Agent} leaves a feedback either on the procedure or the procedure execution. The \texttt{UserQuestion\-Occurrence} allows to keep track of the questions that an \texttt{Agent} may come up with while performing the procedure, and to the possible \texttt{Resource} that allowed to address the question. The \texttt{IssueOccurrence} models the occurrence of an Error that may be encountered by an \texttt{Agent} at some point during execution, and allows to record the cause, as identified by the \texttt{Agent}, and the solution that has been applied by the \texttt{Agent}. Such \texttt{Error} may be associated with an \texttt{errorCode}, and may be addressed by a specific fallback (Multi)Step included in the Procedure. 
An \texttt{Agent}, while interacting with a procedure, can play a certain \texttt{Role} (e.g., editor, supervisor, user). The class \texttt{RoleInTime} makes it possible to model a role that an agent may have and to restrict it to a particular \texttt{PeriodOfTime}.

A \texttt{Procedure} can reference (property \texttt{references}) different \texttt{Resource}s, like documents and images that constitute relevant documentation, and it can be extracted from (\texttt{wasExtractedFrom}) a \texttt{Resource}, as in the case of e.g. a PDF containing some text describing a procedure to execute. 

\paragraph{PKO industry} With this ontology module, it is possible to represent a \texttt{Machine}, or \texttt{Device}, and associate it with its \texttt{MachineType}, \texttt{Location} (e.g., a \texttt{Factory}), and manufacturer (\texttt{wasManufacturedBy}).
Moreover, a \texttt{Step} can be linked to \texttt{Personal\allowbreak{}Pro\allowbreak{}tec\allowbreak{}tive\allowbreak{}Equipment} items that may be needed for safety reasons, e.g., gloves, harness, etc. (\texttt{requiresPPE}).
In the context of LOTO procedures, which are safety procedures used to ensure that dangerous machines are properly shut off during maintenance/repair activities\footnote{LOTO procedures are core to our use cases, provided by Beko Europe.}, a \texttt{Step} is usually linked also to one (or more than one) \texttt{Padlock}, that are locks to be attached to the machine in order to prevent their re-activation. Padlocks can be of different types (e.g., a padlock specific to butterfly valves). Also, this module defines a number of \texttt{EnergySource}s, which are the sources of energy on the machines that need to be identified and isolated during the LOTO procedure (e.g., \texttt{ElectricalEnergy}, \texttt{HydraulicEnergy}, etc.).

\section{PKO Usage Example}\label{sec:howto}

This section exemplifies the intended usage of PKO.
We use one of our three motivating use cases, which revolves around the LOTO procedure, as an example, and show how a subset of the following 9 CQs can be reflected in the data: (i) Which is the target of the procedure?, (ii) Which are the resources related to the procedure?, (iii) By whom is the procedure adopted? (iv) Which are the steps of the procedure?, (v) Which is the next step?, (vi) Which is the expected duration of the step? (vii) Which is the padlock required by the step?, (viii) Which step is being executed, when, and by whom?, (ix) Which questions did an agent ask during execution?.

\begin{figure}[h]
\centering
\includegraphics[width=\linewidth]{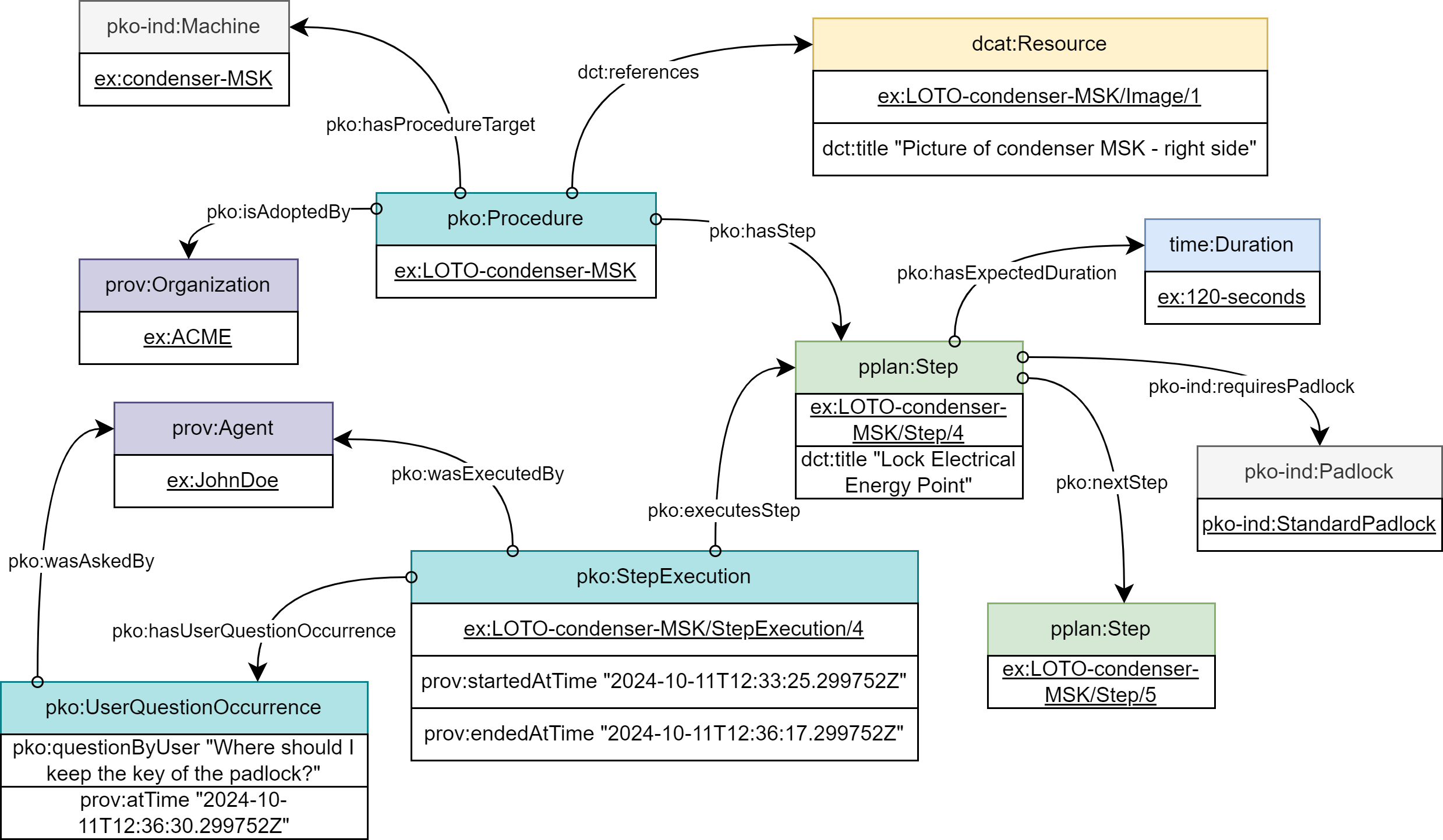}
\caption{Usage example of PKO for LOTO procedures.}
\label{fig:pko-loto}
\vspace{\spaceunderfig}
\end{figure}

Figure \ref{fig:pko-loto} depicts a snapshot of an example LOTO procedure, and its execution, using PKO. The \texttt{ex:LOTO-condenser-MSK}, instance of the class \texttt{pko:Pro\-ce\-dure}, is the LOTO procedure that applies to the condensers (\texttt{pko:hasProce\-dure\-Target}) of the \emph{ACME} \texttt{prov:Organization}. The procedure is also accompanied by some images of the machine to be shut off (\texttt{dct:references} \texttt{dcat:Re\-source}), e.g., the picture of its right side. \texttt{ex:LOTO-condenser-MSK} defines a number of steps to be executed: the fourth \texttt{pplan:Step} in the sequence (\texttt{ex:LOTO\--condenser-MSK/Step/4}) requires to \enquote{Lock Electrical Energy Point} using a \texttt{pko\--ind:StandardPadlock}.
The expected \texttt{time:Duration} of this step is \texttt{ex:120\--sec\-onds}.
The \texttt{pko:StepExecution} of such step started at about 12:33PM of 11 October 2024, and ended at about 12:36PM, thus it actually took one minute longer than expected. \texttt{ex:JohnDoe}, i.e., the \texttt{prov:Agent} that executed such step, before starting the next step, asked to the tool supporting the execution where he should keep the key of the padlock. If recurring in executions of other employees, such question could either be used to improve the description of the LOTO procedure, or included in a set of FAQs.

\begin{figure}[h]
\centering
\includegraphics[width=\linewidth]{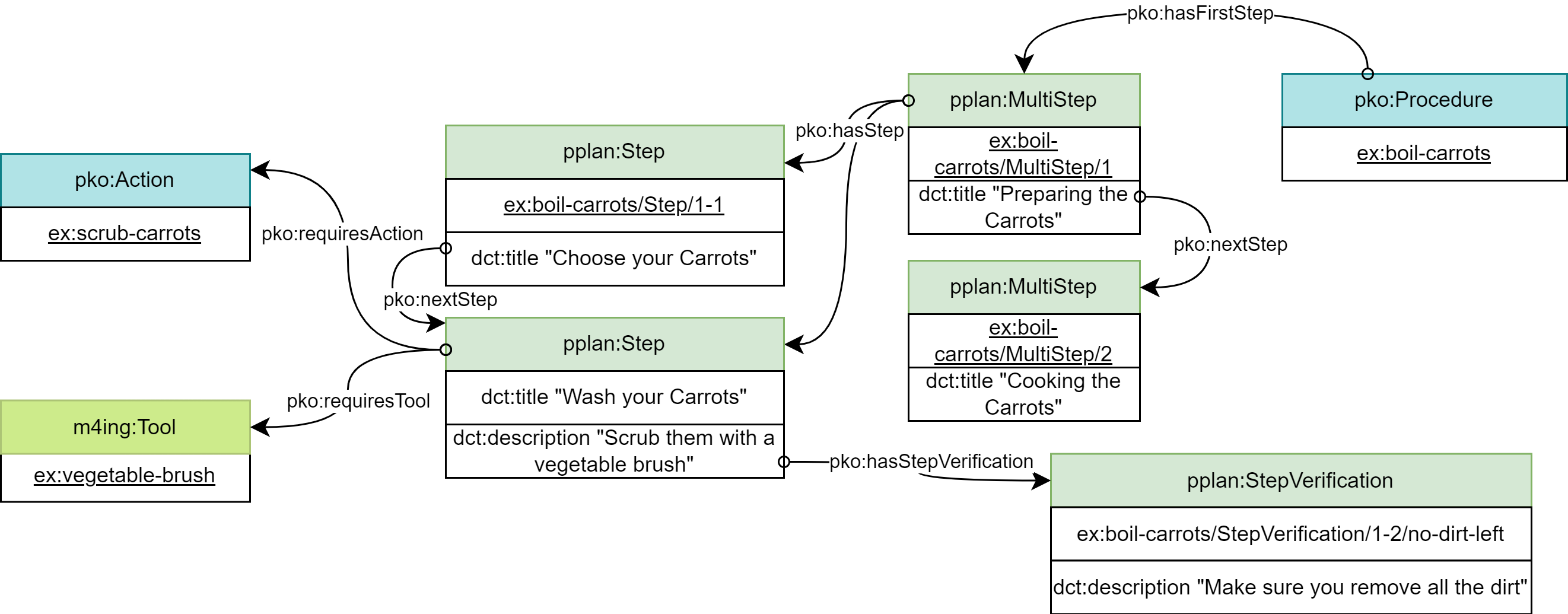}
\caption{Usage example of PKO for recipes.}
\label{fig:pko-recipe}
\vspace{\spaceunderfig}
\end{figure}

To exemplify the potential reuse of PKO in a completely different scenario, we show in Figure \ref{fig:pko-recipe} how PKO can be used to model a recipe, taking the \emph{Boil Carrots} example from WikiHow\footnote{\url{https://www.wikihow.com/Boil-Carrots}}.
The steps of the recipe are grouped into sequential \texttt{pplan:MultiStep}s, that correspond to \emph{parts} in WikiHow. The steps within the multisteps follow a sequential order too (\texttt{pko:nextStep}). The step number \emph{1.2} is also linked to the required \texttt{pko:Action} (\texttt{ex:scrub-carrots}) and the required \texttt{m4ing:Tool} (\texttt{ex:vegetable-scrub}). Moreover, in the description of the step it is recommended to \enquote{make sure you remove all the dirt}: this recommendation can be represented as a way to verify whether the step has been correctly executed (\texttt{pko:StepVerification}).
PKO is easy to extend for specific use cases, e.g., in the context of a recipe, the step could be also annotated with the different ingredients to be used.

\section{PKO Evaluation and potential Impact}\label{sec:evaluation-impact}

\subsection{Evaluation}\label{sec:evaluation}

\emph{Pitfalls} To detect pitfalls in PKO, we used the automatic scanning OOPS! tool \cite{poveda2014oops}. The resulting reports identified no critical issues, but only important and minor ones.
Some pitfalls refer to modelling choices of external ontologies we reuse and we do not have control on, e.g., \texttt{dct:MediaTypeOrExtent} is reported as a class that refers to two different concepts. Most missing annotations arise on ontology entities that we directly reuse from external ontologies, thus their labels/comments are included in their source code, and we use \texttt{skos:scopeNote}
to clarify their usage in PKO. The same applies to warnings about missing inverse relations and domains/ranges: in this case, we used \texttt{dcam:domainIncludes} and \texttt{dcam:rangeIncludes} to annotate external properties with the domain(s) and range(s) we use in PKO.
Instead, we added the reported missing annotations (\texttt{rdfs:comment}) and missing inverse relations to some internal entities, when relevant.
We also looked into the 3 pairs of classes reported as possibly equivalent, and found that none of them could actually be declared as such. 
The scanning of the final version only reports an important pitfall related to our design choice of not specifying disjoint axioms between subclasses. Indeed, we decided to limit the use of logical axioms, while specifying the intended usage of classes and properties through annotations and documentation, and leaving potential validation to the definition of shapes (e.g., SHACL).

We further evaluated PKO using the following well-established criteria for ontology evaluation \cite{vrandevcic2009ontology} as described hereafter.

\emph{Accuracy} The ontology has been developed based on requirements collected and validated by domain experts and relevant users, from three heterogeneous industrial use cases. Moreover, we reused widely used  ontologies.

\emph{Clarity} All terms that we use in the ontology have been validated by our domain experts. Labels, comments and usage notes are provided in order to make the ontology clearer also to external users, supporting its reuse.

\emph{Adaptability} PKO is designed as a modular ontology to support reuse and extensibility. So far, it has been split into two modules, but additional modules may be created.

\emph{Completeness} We checked that all CQs could be translated into SPARQL queries, and no gaps in the vocabulary could be found\footnote{See \url{https://github.com/perks-project/pk-ontology/blob/master/evaluation/ontology-completeness-pko.xlsx}}, thus the ontology can be considered as complete with respect to our current set of requirements. 

\emph{Efficiency} The minor use of logical constraints and the modularity of the ontology ensure its efficiency.

\emph{Conciseness} The collected CQs drove the ontology development, thus the its boundary is well defined. When possible, we reused existing ontologies, and created new entities only when they were missing, making the ontology concise.

\emph{Consistency} No inconsistency was found by the HermiT reasoner v1.4.3.456.

\emph{Organizational fitness} PKO covers the requirements collected by the three different industrial use cases, and the ontology will be used by them in a pilot evaluation using tools for the collection and execution of PK that are based on PKO (cf. Section~\ref{sec:impact}. Moreover, from early interactions with other researchers and industry practitioners, we already discovered potential reuses of PKO to address requirements from other industrial organizations and scenarios. 

\subsection{Impact}\label{sec:impact}


\emph{PK elicitation} 
The first use of PKO is to help domain experts in modeling their specific procedures. To overcome the scarceness of explicit PK, due to the domain experts' lack of time and knowledge engineering skills, we \textcolor{black}{designed and developed} a Web tool that guides step-by-step the user in providing all the needed pieces of information to document a procedure.
Such tool leverages the PKO ontology, presents to the user an easy-to-understand Web form to fill in (whose structure is derived from the classes and properties of the PKO ontology), and transparently transforms user inputs into a KG that complies with the PKO reference ontology. In this way, the domain expert is guided in turning their tacit knowledge into an explicit form; moreover, the same Web form can be used to update a previously entered procedure, in case modifications and updates are required.

The form\footnote{Available here: \url{https://perks-project.github.io/pko-rapid-triples/}} is structured so to guide the domain expert in collecting the relevant information in an orderly manner. 
Once filled in, the tool generates the description of the provided  procedure as RDF described by the PKO ontology, without the need for the operator to understand KG technologies. 
%
Moreover, given that the tool is based on the PKO ontology, all procedures collected through this tool would share the same semantic representation, thus supporting data interoperability across procedures and possibly across industrial companies.

The initial testing of this tool with domain experts gave positive results in terms of simplicity and intuitiveness of the tool, which indeed helps in eliciting PK from tacit knowledge, without requiring to directly deal with the underlying triples. More work may still be needed in order to reduce the PK collection effort, by adding AI-based support during the elicitation phase.

\emph{PK exploitation} 
The second use of PKO is to help industrial operators in correctly executing the modelled procedures. In this case, we adopted a solution based on a chatbot\footnote{The chatbot was developed by Onlim (\url{https://onlim.com/en/}) on their Enterprise Conversational AI platform empowered by KGs.} that can be consulted by the operator in a question-answering conversation to find relevant information about a procedure. This Coversational AI assistant is powered by and leverages a  KG containing procedures modelled according to PKO, to provide relevant answers to the workers' questions.  From a technological point of view, the chatbot is implemented through a KG-empowered RAG pipeline~\cite{xu2024retrieval} that leverages LLMs to handle the natural language interaction with users, still making sure to avoid hallucinations thanks to the retrieval from the PK graph.
The chatbot therefore retrieves the relevant procedure from the KG, and organises the interaction with the user according to the procedure structure, for example displaying  a set of ``cards'' corresponding to the procedure steps, to guide the operator  in the correct execution.



The initial testing of this chatbot with some representatives of the target users also gave positive results, as the Conversational AI guidance appears to be more useful and \enquote{ready-to-use} than consulting a possibly outdated or incomplete documentation. Moreover, the chatbot provides a solution both for novice/inexperienced users, who leverage the step-by-step support, and for more experienced users, who may instead ask specific questions only in case of need.

\emph{Adoption by external parties} We expect a wide potential reuse of PKO by external parties. Indeed, we presented the initial part of our work on PKO at a relevant event on industrial information modeling~\cite{carriero2024towardspko}, and we received multiple expressions of interest from different actors that acknowledged the potential usefulness of using PKO in different scenarios, both in the manufacturing domain and in other industries (such as media, cultural heritage, agriculture). One interesting point that emerged and that we are currently exploring with external collaborators is related to the world of Digital Twins, in which services, capabilities and enablers can be described (at least partially) with PKO and then can be acted upon for simulation and testing purposes~\cite{oakes2024towards}.

\section{PKO Availability and Maintenance}\label{sec:publication}

\emph{Availability}\label{sec:availability}
For all ontology modules, we created permanent URIs with the W3C Permanent Identifier Community Group\footnote{\url{https://w3id.org/}}, which provides a secure URL re-direction service for Web applications.
We used the Widoco\footnote{\url{https://github.com/dgarijo/Widoco}} tool for generating the ontology documentation.
PKO is available on a public GitHub repository\footnote{\url{https://github.com/perks-project/pk-ontology}}. GitHub is also used for version control, since PKO is published with incremental releases, \textcolor{black}{which are associated with dedicated Zenodo DOIs\footnote{\url{https://doi.org/10.5281/zenodo.15007061}}}.
Apart from the ontology code, the repository stores all documentation about the ontology, including the collected requirements.
The maintenance is supported by an issue tracker, which allows to propose additions/changes/corrections/removal directly as GitHub issues\footnote{\url{https://github.com/perks-project/pk-ontology/issues}}. Such proposals are then examined by the PKO ontology engineers and implemented if applicable.
PKO license is Attribution-ShareAlike 4.0 International (CC BY-SA 4.0).
In the future, we will upload the PKO on the Linked Open Vocabulary (LOV), in order to make it more easily reusable.

\emph{Maintenance}\label{sec:maintenance}
PKO will continue to evolve along with the evolution of its requirements. As said above, the piloting of the described solutions in the three motivating use cases may lead to the generation of new requirements, or to the update of already existing ones. Such requirements may result in the extension of the current model and a different modularisation of the ontology. Moreover, the expected potential reuse of PKO in other additional industrial scenarios, as discussed in Section \ref{sec:impact}, shows the relevance and wide applicability of PKO beyond its original boundaries. 
Therefore, PKO long-term sustainability is guaranteed by our commitment to maintain and evolve PKO in the long term.

\section{Conclusions}\label{sec:conclusions}

In this paper, we presented the first version of the Procedural Knowledge ontology (PKO), developed to support the management and reuse of procedural knowledge.
Following the Linked Open Terms methodology, we involved throughout the different steps domain experts and users to collect a comprehensive set of requirements from three different industrial scenarios. 
The PKO ontology reuses existing ontologies, while creating new ontology entities when there was a gap in the state of the art. 
It has been designed as a modular ontology to support reuse and extensibility. So far, it has been split into two modules (\emph{core} and \emph{industry}).

PKO – published online according to the Semantic Web best practices and FAIR principles – underwent a thorough evaluation, both manual and with an automatic diagnosis tools, and it guarantees accuracy, adaptability, clarity and completeness. 
We reported two main usage examples, and discussed the impact resulting from its adoption for PK elicitation and reuse.

In the next period, we aim at validating the proposed ontology by piloting its use through tools that support both elicitation and reuse of procedures in the three industrial use cases that originated PKO its requirements. Moreover, we will investigate and work on possible extensions, e.g., in the Digital Twins scenario, to show its further applicability to diverse industrial use cases.
From a technical point of view, to further facilitate PKO reuse, we will generate a set of SHACL shapes to support the validation of PKO-based KGs.


\begin{credits}
\subsubsection{\ackname} This work is partially supported by the PERKS project, co-funded by the European Commission (Grant id 101120323). We would like to specifically thank the industrial domain experts from Beko Europe, Fagor Automation and Siemens who provided their procedure scenarios and requirements for PKO.
\end{credits}
%
%
%
 \bibliographystyle{splncs04}
 \bibliography{references}
%




\end{document}